\documentclass[10pt,twocolumn,letterpaper]{article}

\usepackage{cvpr}
\usepackage{times}
\usepackage{epsfig}
\usepackage{graphicx}
\usepackage{subfig}
\usepackage{amsmath}
\usepackage{amssymb}
\usepackage{amsthm}
\usepackage{xcolor}
\usepackage{multirow}

\usepackage[pagebackref=true,breaklinks=true,letterpaper=true,colorlinks,bookmarks=false]{hyperref}

\cvprfinalcopy % *** Uncomment this line for the final submission

\def\cvprPaperID{3260} % *** Enter the CVPR Paper ID here

\newcommand{\dataset}{Shading Annotations in the Wild}
\newcommand{\datasetshort}{SAW}

\newcommand{\norm}[1]{\left\lVert#1\right\rVert}

\newcommand{\smooth}{{\sc S}}
\newcommand{\ns}{{\sc NS}}
\newcommand{\nssb}{{\sc NS-SB}}
\newcommand{\nsnd}{{\sc NS-ND}}

\newcommand{\rie}{{\mathcal{R}}}
\newcommand{\sie}{{\mathcal{S}}}
\newcommand{\ri}{{$\mathcal{R}$}}
\newcommand{\si}{{$\mathcal{S}$}}

\newenvironment{packed_item}{
\begin{itemize}
  \setlength{\itemsep}{1pt}
  \setlength{\parskip}{2pt}
  \setlength{\parsep}{0pt}
}{\end{itemize}}

\ifcvprfinal\fi
\begin{document}

%%%%%%%%% TITLE
\title{Shading Annotations in the Wild} % Replace with your title

\author{Balazs Kovacs \ \ \ \ \ \ \ \ \ \ \ \ \ \ \  Sean Bell \ \ \ \ \ \ \ \ \ \ \ \ \ \ \ Noah Snavely \ \ \ \ \ \ \ \ \ \ \ \ \ \ \  Kavita Bala\\
{Cornell University}\\
}

\maketitle

%%%%%%%%% ABSTRACT
\begin{abstract}
Understanding shading effects in images is critical for a variety of
vision and graphics problems, including intrinsic image decomposition,
shadow removal, image relighting, and inverse rendering. As is the
case with other vision tasks, machine learning is a promising approach
to understanding shading---but there is little ground truth shading
data available for real-world images.  We introduce \dataset{}
(\datasetshort{}), a new large-scale, public dataset of shading annotations in
indoor scenes, comprised of multiple forms of shading judgments
obtained via crowdsourcing, along with shading annotations
automatically generated from RGB-D imagery. We use this data to train
a convolutional neural network to predict per-pixel shading
information in an image. We demonstrate the value of our data and
network in an application to intrinsic images, where we can reduce
decomposition artifacts produced by existing algorithms.
Our database is available at
\href{http://opensurfaces.cs.cornell.edu/saw/}{http://opensurfaces.cs.cornell.edu/saw}.

\end{abstract}

%%%%%%%%% BODY TEXT
\section{Introduction}

Understanding images requires reasoning about the shapes and materials
in scenes, where the appearance of objects is modulated by
illumination. A large body of research in scene understanding has
focused on shape and materials, with lighting often overlooked or
discounted as a nuisance factor.
%%%%%%
However, understanding shading and illumination in images is critical
for a variety of problems in vision and graphics, including intrinsic
image decomposition, shadow detection and removal, image relighting,
and inverse rendering. How can we make progress on understanding
illumination in natural images? As with other problem domains, we
believe that data is key.
%%%%%%
Large-scale datasets
such as ImageNet~\cite{deng2009imagenet},
COCO~\cite{lin2014microsoft}, Places~\cite{zhou2014places},
and MINC~\cite{bell15minc}
have had significant impact in advancing research in object detection,
scene classification and understanding, and material recognition.
This success motivates the creation of a similar dataset for shading
information.

\begin{figure}[t]
\centering
  \begin{tabular}{cccc}
    \includegraphics[width=0.45\linewidth]{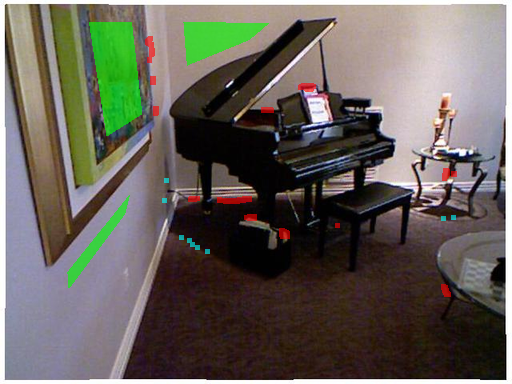} &
    \includegraphics[width=0.45\linewidth]{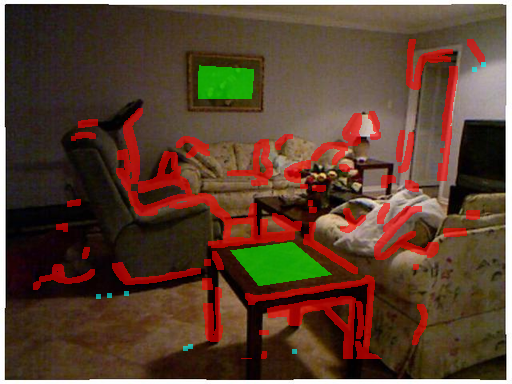} \\
    \includegraphics[width=0.45\linewidth]{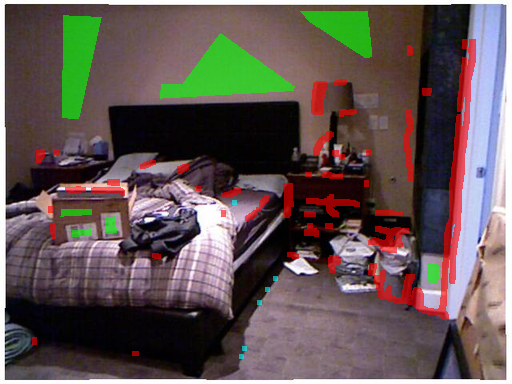} &
    \includegraphics[width=0.45\linewidth]{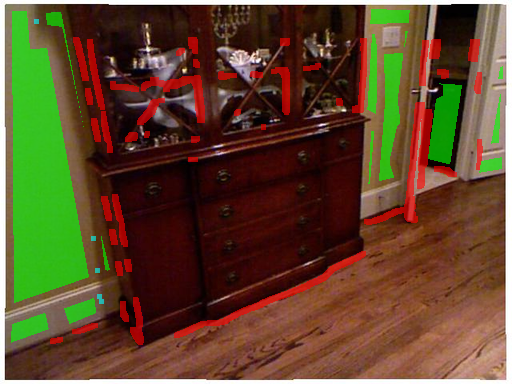}
  \end{tabular}
  \caption{{\em Examples of annotations in the \datasetshort{} dataset.}
\textcolor{green}{{\bf Green}} indicates regions of near-constant shading (but
with possibly varying reflectance). \textcolor{red}{{\bf Red}} indicates edges
due to discontinuities in shape (surface normal or depth). \textcolor{cyan}{{\bf
Cyan}} indicates edges due to discontinuities in illumination (cast shadows).
Using these annotations, we can learn to classify regions of an image into
	different shading categories.}
\label{fig:annotations}
\end{figure}

In this paper, we present a new, large-scale crowdsourced dataset of
{\em Shading Annotations in the Wild} (\datasetshort{}). An important challenge
in constructing a dataset of shading information is deciding what manner of
shading information to collect, and how to collect it. We consider several
possible approaches to collecting such data, and note that a key subproblem
across many tasks is to determine whether an image edge is due to variation in
reflectance, illumination, or some other cause (as with the Retinex algorithm
for intrinsic images~\cite{land1971}). This observation leads us to collect two
types of shading annotations in a large set of images: (1) image regions of
approximately constant shading, and (2) examples of discontinuities in
illumination (i.e., cast shadow boundaries), or shape (e.g., depth or surface
normal discontinuities). These kinds of annotations are illustrated in
Figure~\ref{fig:annotations}. We show how to collect these annotations at scale
using a combination of crowdsourcing and automatic processing. Our dataset
includes  15K shadow boundary points and 24K constant shading regions from
nearly 7K photos.

Using our new dataset, we train a convolutional neural network (CNN)
to identify various types of shading in new images, and demonstrate
competitive performance in this shading classification task compared
to a number of baselines. Finally, we demonstrate the value of our
data and learned network in an application to intrinsic image
decomposition, where we can reduce mistakes commonly made by existing
algorithms, namely, when texture due to reflectance is incorrectly
attributed to shading.

In summary, our contributions are:
\begin{packed_item}
\item a new large-scale dataset of shading annotations collected via
  crowdsourcing,
\item a CNN trained to recognize shading effects using this data,
  and a comparison to baseline methods, and
\item an example use of this model as a smooth shading prior to improve
  intrinsic image decomposition.
\end{packed_item}

\section{Related Work}\label{sec:related}

Our goal is to build a dataset specifically addressing shading in
images, and large enough to be well suited for machine learning. There
exist a number of related datasets, but to our knowledge, none achieve
both of these goals.

\medskip
\noindent{\bf Intrinsic images.} Intrinsic image decomposition is a classic,
ill-posed problem involving separating an image into the product of a
reflectance and a shading layer. Grosse
\etal~\cite{grosse2009} introduced the MIT Intrinsic Images dataset,
containing 16 objects with ground truth reflectance and shading. This dataset
has led to important progress in intrinsic image decomposition, but the small
size of the dataset, and its focus on single objects rather than entire scenes,
means that it is not well suited to machine learning approaches on natural
images. Beigpour \etal capture a dataset of similar size,
but with multiple illuminants~\cite{Beigpour_2015_ICCV}.
%%%%%%%
Bell~\etal~released the Intrinsic Images in the Wild
(IIW) dataset~\cite{bell2014intrinsic}, a large-scale dataset with
over 5K real-world indoor photos, with relative reflectance judgments
between millions of pairs of points.
%%%%
However, IIW only contains information about reflectance, and thus
only captures indirect information about shading. As a result,
intrinsic image algorithms evaluated on IIW data can sometimes
shuffle error into the shading channel without
penalty.  Finally,
synthetic datasets (from rendered CG scenes) also provide a way to
obtain ground truth shading for intrinsic images and other
problems~\cite{Butler:ECCV:2012,beigpour2013intrinsic,BKPB17}. However, we
find that synthetic scenes still cannot fully represent the complexity of
natural images.

\medskip
\noindent{\bf Depth datasets.}  Several datasets contain RGB-D (depth)
data,
including NYUv2~\cite{Silberman:ECCV12}, SUN RGB-D~\cite{song2015sun}, and many
others~\cite{firman-cvprw-2016}. These datasets can be used to train
algorithms to predict depth or surface normals from a single
image~\cite{zeisl2014discriminatively,eigen2015predicting,li2015depth,Bansal16,chen2016single}. These
shape cues (particularly surface normals) are related to shading, but
do not capture critical illumination effects such as cast
shadows. Hence, we draw on RGB-D data to augment our dataset, but use
crowdsourcing to annotate additional shading information.

\medskip
\noindent{\bf Other illumination datasets.}
Other datasets capture particular types of illumination information,
such as sun direction~\cite{lalonde2009estimating}, environment
maps~\cite{lalonde2014lighting}, or shadows in outdoor
scenes~\cite{zhu2010learning,lalonde2010detecting}. These datasets tend
to focus exclusively on outdoor illumination (e.g., from the sun), or
only support a particular task (e.g., hard shadow detection
and removal~\cite{lalonde2010detecting,guo2011shadow}). Others have
presented algorithms for estimating illumination from images,
e.g., for object insertion tasks~\cite{karsch2011rendering},
relighting~\cite{haber2009relighting,ren2015image}, or more general
inverse rendering problems~\cite{yu-SIGGRAPH1999}. However, these
generally require user input or multiple images. One of our goals is
to help advance such methods for illumination modeling and editing by
providing data for use in machine learning methods.

\section{\dataset}\label{sec:dataset}

Our goal is to create an extensive
dataset of shading phenomena in indoor
scenes. Ideally we would collect per-pixel, dense absolute shading measurements
for each image, as with the MIT dataset~\cite{grosse2009}.  Unfortunately, the
gray spray painting method they used is not feasible for whole indoor scenes.
Synthetic scenes are a potential alternative to provide dense ground truth, but
we found that it is difficult to build a large enough dataset of synthetic
images that can fully represent the complex illumination in the real world.

Bell~\etal targeted a broad set of real-world scenes
by annotating Flickr images in their Intrinsic Images in the Wild dataset~\cite{bell2014intrinsic}.
They argued that while humans cannot provide absolute reflectance or shading values,
they can disentangle reflectance from shading by making
pairwise reflectance judgments.  Reflectance values tend to be sparse
in indoor scenes, due to the overwhelming presence of human-made objects, which
is often used as a prior in the intrinsic image
literature~\cite{gehler-NIPS2011,bell2014intrinsic,zhou2015learning}.
Conversely, this sparsity observation does not hold for shading, which is often
smooth and varies over a wide intensity range in natural scenes. Bell \etal
pointed out that this makes it harder for humans to make relative shading
judgments between arbitrary point pairs in images, so they did not collect
pairwise shading annotations.

Our contribution is to identify and collect useful shading annotations that
human beings can provide in a crowdsourced setting, at scale and with high
accuracy.

\subsection{Images}
To create a comprehensive dataset of shading phenomena, we chose to build on
the Intrinsic Images in the Wild (IIW) dataset~\cite{bell2014intrinsic} which
has complementary data on relative reflectance annotations for 5,230
images.\footnote{\datasetshort{} images are a superset of IIW images except for
two images (IDs: 24541, 24851), which are atypical photos that we exclude.  One
is a painting, and the other is a closeup of a book cover.} We further added
1,449 images with RGB-D data from the NYU Depth Dataset
v2~\cite{Silberman:ECCV12} to have images from which we can get ground truth
depth and surfaces normals. In total, the SAW dataset has 6,677 images.

\begin{figure}[t]
  \centering
  \begin{tabular}{ccc}
    \includegraphics[width=0.3\linewidth,height=0.3\linewidth]{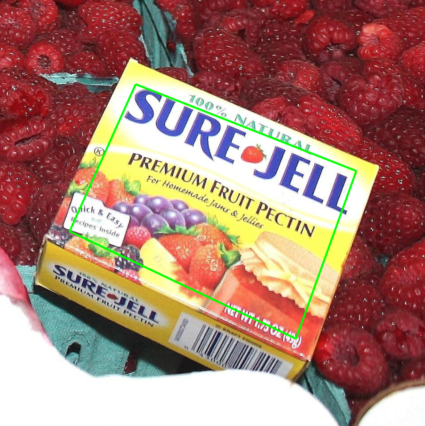} &
    \includegraphics[width=0.3\linewidth,height=0.3\linewidth]{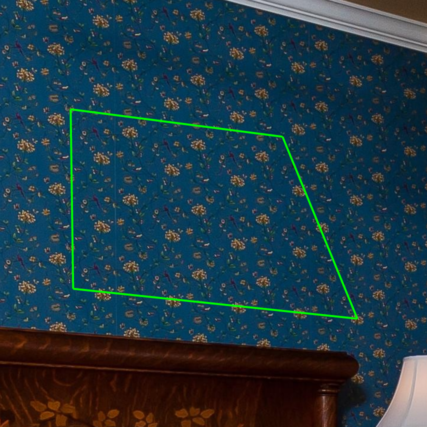} &
    \includegraphics[width=0.3\linewidth,height=0.3\linewidth]{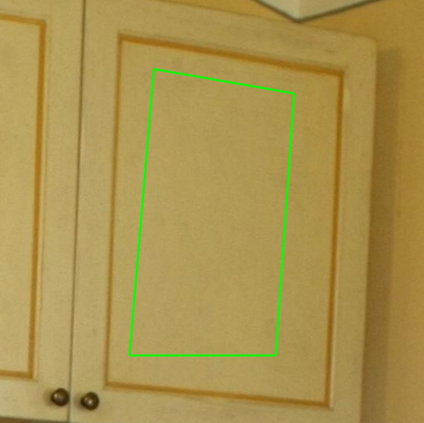}\\
    \includegraphics[width=0.3\linewidth,height=0.3\linewidth]{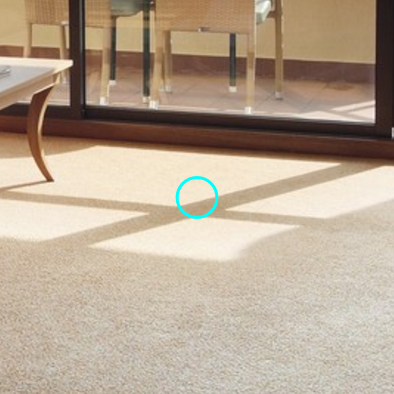} &
    \includegraphics[width=0.3\linewidth,height=0.3\linewidth]{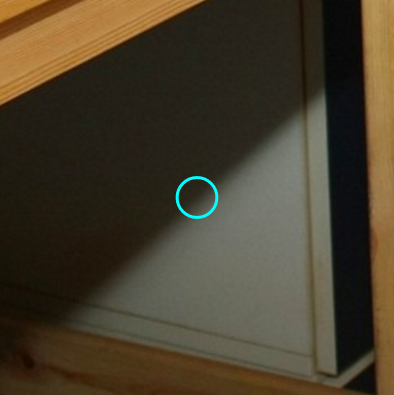} &
    \includegraphics[width=0.3\linewidth,height=0.3\linewidth]{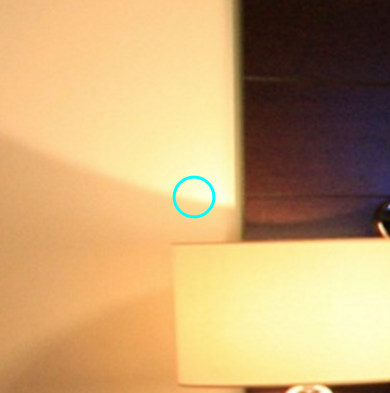}
  \end{tabular}
  \caption{{\em Our shading annotations.} {\bf First row:} Constant
	shading regions \smooth{} (green polygons). {\bf Second row:} Shadow
	boundary annotations \nssb{} (cyan circles). The constant shading regions
    span the range from textured to textureless (the average
    color gradient magnitude for the regions shown from left to right is
    $3.972$, $0.295$, and $0.1$).
  }
\label{fig:shading-annos-closeup}
\end{figure}

\subsection{Shading annotation taxonomy}\label{subsec:shading-taxonomy}
Our goal is to collect shading annotations at scale. We taxonomize shading
into two types: smooth (\smooth{}) and non-smooth (\ns{}),  where the
non-smooth shading is further split into two categories, shadow boundaries
(\nssb{}), and normal/depth discontinuities (\nsnd{}).  Using a judicious
combination of crowdsoucing when needed, and automatic image/scene processing
when possible, our dataset includes these three types of shading annotations.

\subsection{Our annotation pipeline}

\noindent{\bf Pilot study.} Inspired by IIW~\cite{bell2014intrinsic}, our
first attempt to collect shading annotations was to use the same kind of
pairwise comparisons as in IIW, where workers were asked to make a
series of pairwise reflectance judgments. For IIW, workers were shown a
pair of points {\bf 1} and {\bf 2} in an image, and asked to specify whether: (1)
{\bf 1} had a
darker surface color compared to {\bf 2}, (2) {\bf 1} had a brighter surface color compared
to {\bf 2}, or (E) {\bf 1} and {\bf 2} had approximately equal surface brightness (i.e.,
less-than, greater-than, or equal-to judgments). In our case, rather than
collect pairwise reflectance comparisons, our aim was to collect pairwise {\em
shading} annotations.

Reasoning about how lightfields differ between arbitrary points is not easy for
humans~\cite{Ostrovsky05}. Indeed, in this case humans often have to make judgments
over very different regions of an image, and over different materials and
shapes. Hence, we decided to allow workers to pick the points themselves instead
of using the original point pairs from IIW. We created two tasks which ask
workers to ``pick two points with equal shading'' and ``pick two points with
non-equal shading''. Unfortunately, workers struggled with giving us good
quality data for the former task.  Learning from this pilot study we instead
developed a new crowdsourcing pipeline to collect shading annotations that
workers can confidently respond to.

\begin{figure}[t]
  \centering
  \includegraphics[width=\linewidth]{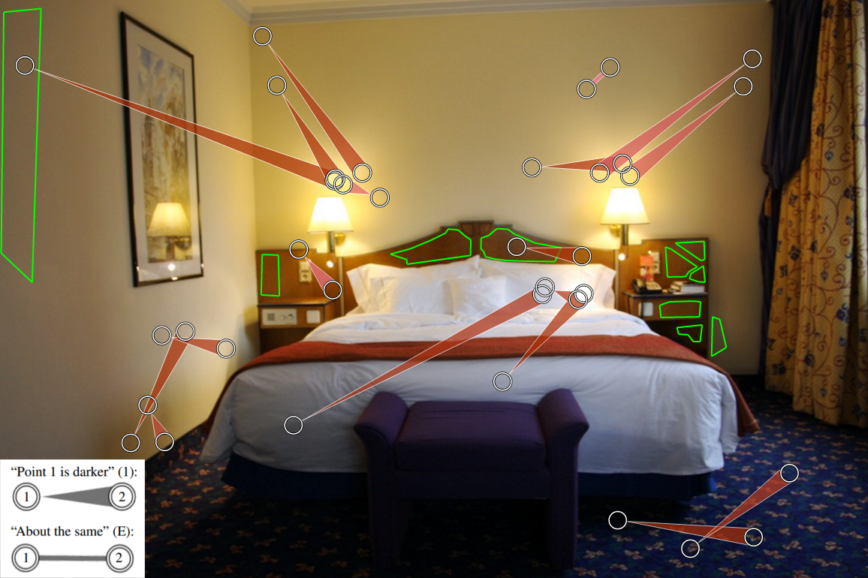}
  \caption{{Two types of shading annotations.}
    {\bf (a)} Constant shading regions (green polygons).
    {\bf (b)} Shading point comparisons (red edges).
    Darker red indicates more confident judgments.
}
\label{fig:shading-annos-whole}
\end{figure}

\medskip
\noindent{\bf Collecting \smooth{} annotations.}
Knowing that human beings have difficulty reasoning about distant shading, we
ask workers to instead annotate local regions which they select to have
approximately constant shading. Since shading tends to be smooth in small
regions, they can do this task reliably.  Further, we get much more
data from a region annotation than a pairwise comparison between two points.
Thus, we were able to collect \smooth{} annotations at scale with a small
number of selected workers.
These \smooth{} annotations are collected over both IIW
and the NYU dataset.  See Figure~\ref{fig:shading-annos-closeup} for examples,
and Section~\ref{sec:S} for details.

\medskip
\noindent{\bf Collecting \ns{} annotations.}  Non-smooth shading
arises from a variety of causes, such as shadow boundaries or changes
in the shape of a surface (e.g., through depth discontinuities or
normal discontinuities). We employ a combination of automated
scene/image processing and crowdsourcing to collect these annotations.

First, we note that shape discontinuities (i.e., depth or normal changes) can
be obtained from existing datasets like the NYU RGB-D dataset. Therefore, instead
of crowdsourcing these annotations (\nsnd{}), we automatically generate
normal/depth discontinuities from the ground truth RGB-D data. More details are
provided in Section~\ref{sec:NSND}.

Another type of non-smooth shading arises at shadow boundaries.  For each
image in IIW and the NYU dataset, we allow workers to select point pairs with
different shading.  Since workers control the pair selection, they are able to
choose cases where they can make a confident decision.  We find that the pairs
of selected points are often on opposite sides of sharp shadow boundaries.
We use this knowledge to automatically generate candidate shadow boundary
points from the pairwise data from workers, which we filter through another
crowd-sourcing task to separate out true shadow boundaries (\nssb{}) from shape
discontinuities.
See Figure~\ref{fig:shading-annos-closeup} for examples, and Section~\ref{sec:NSSB} for details.

\begin{figure}[t]
  \centering
  \begin{tabular}{cc}
    \includegraphics[width=0.46\columnwidth]{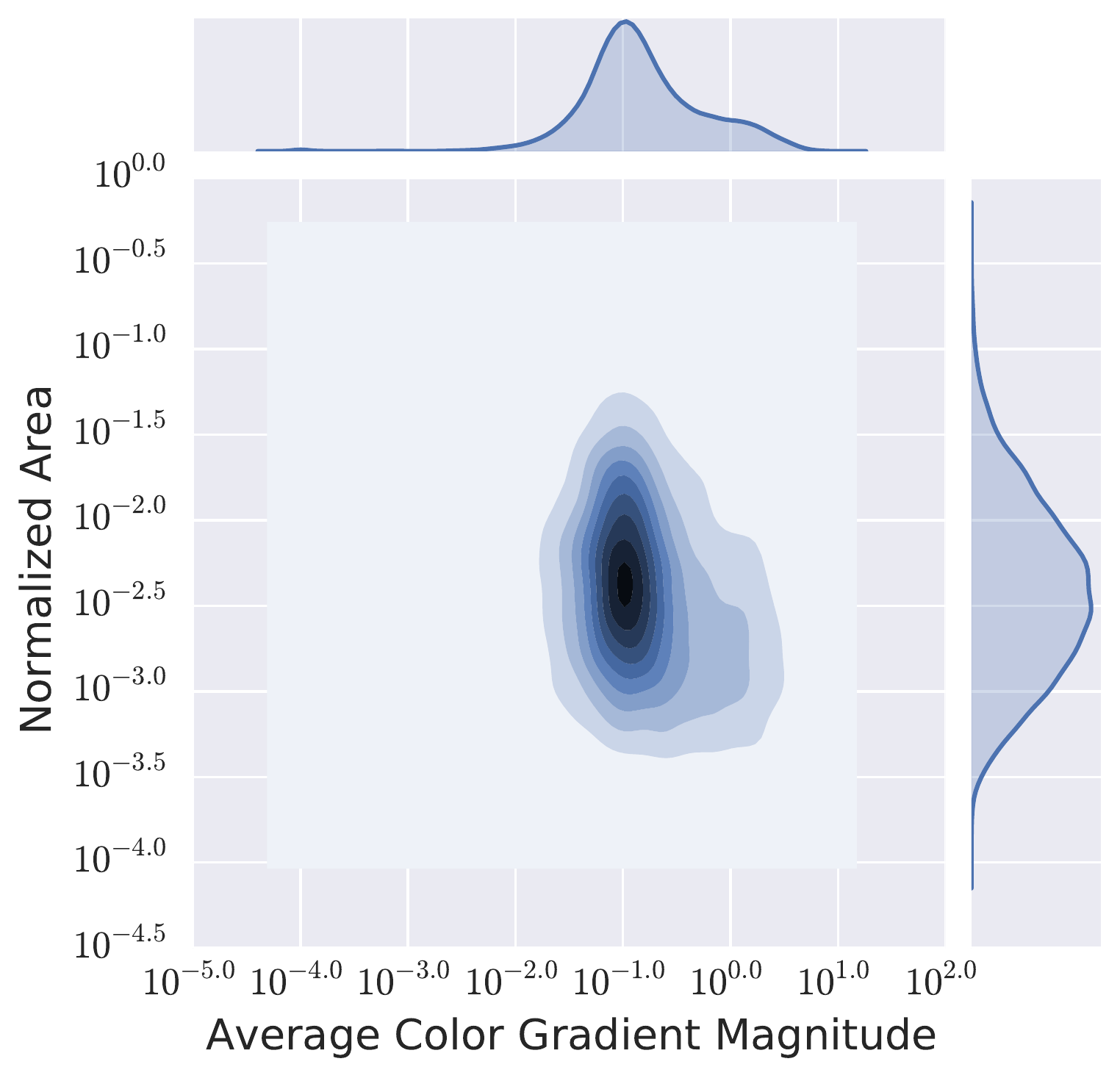}&
    \includegraphics[width=0.46\columnwidth]{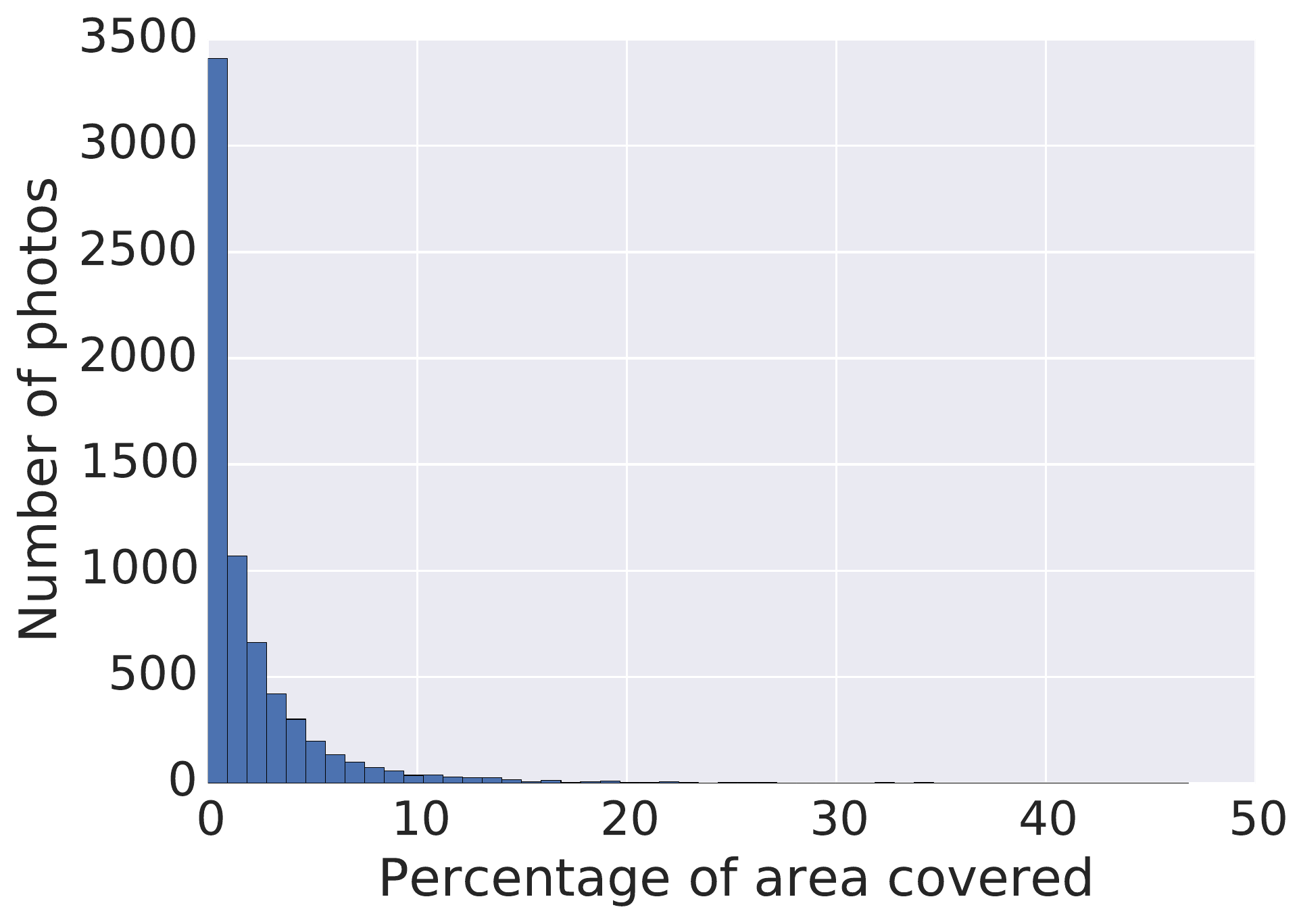}
  \end{tabular}
  \caption{{\em Statistics of constant shading regions.}
    {\bf Left:}
    Joint plot of the log average color gradient magnitude over each constant
    shading region and
    the log normalized area (1 means that the region covers the entire
    image). The gradient magnitude is correlated with how textured the region
    is. Textured regions are valuable because constant shading cannot be easily
    predicted based on simple pixel intensity measurements.
    {\bf Right:}
    Histogram of the percentage of total image area covered by constant shading regions. As expected, most of the regions are
    relatively small, since it is unlikely that shading is constant over large
    regions. See the supplemental material for more analysis.}
\label{fig:data-stats}
\end{figure}

\subsection{Collecting \smooth{} annotations}
\label{sec:S}

For this task, each worker was asked to draw a polygon around an area which has approximately
constant shading. The notion of constant shading is hard to understand for most
workers, so to guide workers to submit higher quality regions we added extra
criteria: the region has to be flat/smooth, opaque (i.e., non-transparent),
non-glossy, and non-bumpy (i.e., have no surface normal variation).  Based on our
pilot study, we further restricted the region to be composed of a single
type of material (e.g.\, wood, plastic), and not fabric, which tends to have
small bumps in most situations. However, we explicitly request that workers
annotate textured regions when possible, so that we do not simply collect
regions with a single dominant color (e.g., painted walls).  Such textured
regions are very valuable, because constant shading cannot be easily predicted
from simple pixel intensity measurements in these regions. We allowed eight
MTurk workers who previously provided high-quality submissions to work on this
task.

For quality control, we additionally sent each marked region
through three filtering tasks to address common mistakes.
These tasks asked workers to (1) ``click on
flat/smooth regions with one material type'', (2) ``click on glossy regions'',
and (3) ``click on regions which have varying shading''.  Since these tasks are
much simpler than the first task, we did
not need to hand-select workers here.  For each smooth shading
region, we collected five responses for each of the three tasks and used
CUBAM~\cite{welinder-NIPS2010} to aggregate the votes into a single decision.
We kept regions that passed all tests (i.e., regions that were
flat/smooth with one material type, non-glossy, and did not exhibit varying
shading).

In total, we collected 23,947 smooth regions (\smooth{}), which
on average covered 0.6\% of the image area. The cost of this
task was \$0.011 on average for the shading regions, plus \$0.056 for the three quality
control tasks.
Figure~\ref{fig:shading-annos-closeup} (top) shows examples of annotated smooth
shading regions and green polygons in Figure~\ref{fig:shading-annos-whole} show
these regions in the context of an entire scene.
Figure~\ref{fig:data-stats} provides insights into the quality of
the constant shading region data.

\subsection{Collecting \ns{} annotations}
Next, we turned our attention to non-smooth shading annotations (\ns{}). Here we found
from our pilot study that if workers are given a choice of where to position
a pair of points, they can successfully decide which point has darker vs.\
brighter shading.  However, these shading changes could be attributed to both shape
changes (normal/depth discontinuities) or due to shadow boundaries. While we
could crowdsource both these kinds of annotations, the shape discontinuties can
be obtained directly from existing datasets. So we automatically generate \nsnd{}
annotations, and only use crowdsourcing for the \nssb{} annotations.

\medskip
\noindent{\bf Auto-generated NS-ND annotations.}
\label{subsec:autogen}
\label{sec:NSND}
At normal/depth discontinuities, shading tends to be non-smooth. We
generate \nsnd{} annotations using depth maps of scenes from existing datasets
such as NYU Depth Dataset v2~\cite{Silberman:ECCV12}, and normal maps computed from
these depth maps from~\cite{zeisl2014discriminatively}. Given a depth $D$ and
normal map $N$ and thresholds $\tau_{depth}$ and $\tau_{normal}$, we annotate a
pixel $p$ as having non-smooth shading if $\left(\norm{\nabla D}_2\right)_p >
\tau_{depth}$ or $\left(\norm{\nabla N}_2\right)_p > \tau_{normal}$.

We ignore pixels where the Kinect camera used to capture the RGB-D imagery
provides unreliable depths, using masks provided by~\cite{eigen2015predicting}.
We noticed that in some cases, these masks do not sufficiently remove all
incorrect normal/depth regions, and so we use binary erosion with 3 iterations
on each mask and also ignore pixels close to the image boundaries (within 5\%
of the image width).

\begin{figure}[t]
\centering
\includegraphics[width=\linewidth]{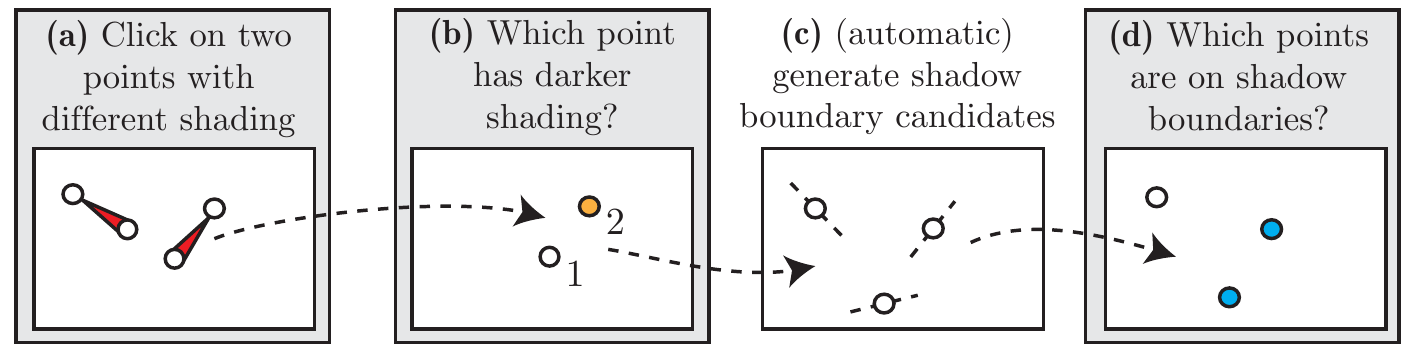}
\caption{{\em Point annotation pipeline.} {\bf (a)} Workers
  are asked to click on two points such that the first has darker shading than the
  second.  {\bf (b)} Then, 5 workers are asked to pick the point with darker shading
  for each point pair. {\bf (c)} Next, we automatically generate a candidate
  shadow boundary point for each point pair based on image gradient. {\bf (d)}
Finally, workers are asked to select shadow boundary points.}
\label{fig:point-pipeline}
\end{figure}

\medskip
\noindent{\bf Crowdsourcing NS-SB annotations.}
\label{sec:NSSB}
Finally, we crowdsource non-smooth shading annotations, with a pipeline focusing
on shadow boundaries (Figure~\ref{fig:point-pipeline}).  The first task in the
pipeline asks workers to select two points such that the first has darker
shading than the second point.  After filtering out comparisons which have
non-opaque or glossy points, we collected five votes for each comparison asking
which point has darker shading (Figure~\ref{fig:point-pipeline}(b), similar
to~\cite{bell2014intrinsic} for relative reflectance judgments).  The original
pair of two points counts as an additional vote for a total of 6 votes.  See the
supplementary for more details. We collected 97,294 shading comparisons with an
average cost of \$0.026. Red edges in Figure~\ref{fig:shading-annos-whole} show
example relative shading judgments.

The last step is to generate and validate shadow boundary points
(Figures~\ref{fig:point-pipeline}(c) and (d)).
Given the shading comparisons, we generate candidate shadow boundary points
for each non-equal shading comparison by finding the point with the highest log
intensity gradient magnitude on the line segment connecting the two points of
the comparison (Figure~\ref{fig:point-pipeline}(c)).
We discard candidate points where the line segment is longer than $0.2$ in normalized
image coordinates, because these point pairs are too far apart and the candidate
point usually lies on a shape discontinuity; or where the maximum gradient magnitude
is smaller than $0.3$, because such intensity differences are hard to notice.
Then we asked five workers if the candidate point is on a shadow boundary (Figure~\ref{fig:point-pipeline}(d)). We
define the term ``shadow boundary'' here to exclude normal or depth
discontinuities. This ensures that we can make a distinction between the
automatically generated normal/depth discontinuity labels (\nsnd{}) from
Section~\ref{sec:NSND} and shadow boundary labels (\nssb{}). We chose the final
shadow boundaries with majority voting.

Using this pipeline, we obtain 15,407 shadow boundary points at an average
cost of \$0.039. Figure~\ref{fig:shading-annos-closeup}
(bottom) shows examples of shadow boundary annotations.
We provide statistics of the collected shadow boundary points in the supplemental material.

\medskip
\noindent{\bf Quality control for crowdsourcing S and NS-SB.} It is important to
control quality when collecting crowdsourcing data~\cite{Allahbakhsh:2013:QCC:2498344.2498595,Gingold:2012:MPH:2231816.2231817}.
Many workers misunderstand instructions or do not read them in detail.
Therefore we implemented tutorials for most of our crowdsourcing tasks and did
not let workers submit data until they passed the tutorial. We also ask
multiple workers the same question and decide the final label by majority
voting or CUBAM~\cite{welinder-NIPS2010}. Finally, we employ sentinels
(questions with known ground truth) to filter out workers with low
accuracy.

\section{Learning to Predict Shading Features}
\label{sec:heatmap}
We demonstrate
the utility of our shading annotation data by training a CNN to make per-pixel
predictions of different types of shading features. We formulate this
problem as classifying each pixel of an image into one of three
classes based on the taxonomy defined in Section~\ref{subsec:shading-taxonomy}: smooth
shading (\smooth{}), normal/depth discontinuity (\nsnd{}), and shadow boundary
(\nssb{}).

\subsection{Dataset processing}\label{subsec:labelgen}
Before we train a classifier, we first convert our dataset into a pixel
labeling for each image (note that only some pixels will be labeled, since our
annotations only partially cover each image). First, we resize all images
such that the maximum image dimension is 512.
Next, we generate smooth shading (\smooth{}) labels from our constant shading
regions
by taking the regions in the resized images and performing binary
erosion with 3 iterations, to reduce the effect of any errors where
constant-shading-region boundaries may touch shadow boundaries. This gives us
25,690,392 smooth shading pixel labels across the entire dataset.

We then generate the normal/depth discontinuity non-smooth shading
(\nsnd{}) labels based on the resized normal/depth maps of the 1,449 NYUv2 images
as described in Section~\ref{subsec:autogen} with $\tau_{normal} = 1.5$ and
$\tau_{depth} = 2.0$. We manually chose the smallest thresholds where we deemed the
annotations to be of high quality.
Finally, we use our shadow boundary point annotations to generate the rest of
the non-smooth shading (\nssb{}) labels.

Note that we perform ``label dilation'' on the non-smooth shading labels when
generating the training set: that is, we also label pixels that are very close
to these non-smooth pixels within a $5 \times 5$ neighborhood.  We do this 
to train a more conservative classifier which does not predict smooth
shading very close to non-smooth shading effects. For the validation and test
set, we do not perform this dilation.
This way for the training, validation, and
test set respectively,
we get 4,758,500/1,512,257/2,418,490
\nsnd{} and 224,886/ 2,107/ 4,267 \nssb{} labels.

\subsection{Network architecture}
We extend Bansal \etal~\cite{Bansal16}'s convolutional neural network (CNN)
architecture for surface normal prediction to learn to predict shading effects
in images using the Caffe deep learning framework~\cite{jia2014caffe}. We use
the same architecture, but change the last fully-connected layer to predict the
three classes described above.

\subsection{Training}
We assign each photo to the training/validation/test sets as follows:
For photos in the original IIW set, we keep the training/test split
used by~\cite{zhou2015learning} and add an additional
training/validation split over their training set. For NYU images, we
use the splits from~\cite{Bansal16}. This gives us 4,142 training, 836
validation and 1,699 test photos.

Since our training data is limited, we initialize the
weights using the normal prediction net of~\cite{Bansal16}, fix the weights of
the convolutional part and only fine-tune the last three fully-connected layers.
We experimented with fine-tuning all layers or only the last fully-connected layer, but observed worse validation performance.
To avoid training bias, it is important to balance the training data across the
three classes. We use a 2:1:1 balancing ratio (\smooth{} : \nsnd{} :
\nssb{}) in our experiments, equivalent to a 1:1 balance between
smooth and non-smooth categories.

Similar to~\cite{Bansal16}, we resize each input image to $224 \times 224$
before passing it to the network, and upsample the output of the
network back to the resolution of the original input image. Since not all
pixels in an image are labeled and we want to enforce class balance, after
passing all images in a batch through the convolutional layers, we randomly
sample pixels for each class over the whole batch according to our balance
ratio, and propagate features only for these sampled pixels to the rest of the
network. Please see the supplemental material for detailed training
parameters.

\subsection{Inference}~\label{subsec:inference}
\begin{figure}[t]
  \centering
  \begin{tabular}{cc}
    \includegraphics[width=0.4\linewidth]{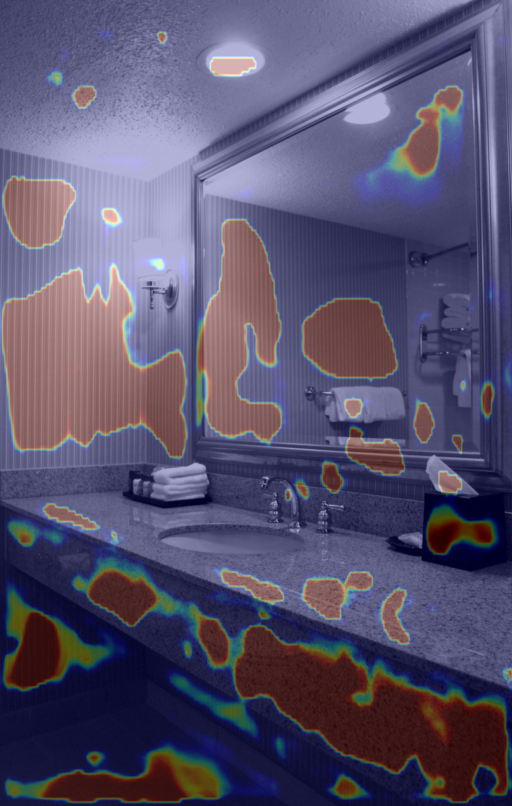} &
    \includegraphics[width=0.3535\linewidth]{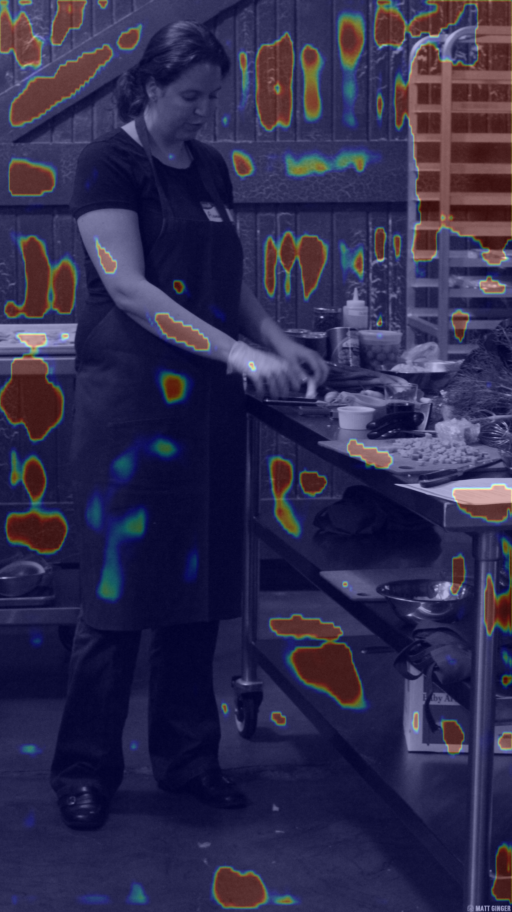}
  \end{tabular}
\caption{{\em Heatmaps for the predicted probability of the smooth shading
  class (\smooth{}) overlaid on the original input images.} All images are selected from the
test set. See Section~\ref{subsec:inference} for discussion of these results.}
\label{fig:heatmaps}
\end{figure}

At inference time, we are interested in predicting the probability of the
shading being smooth for each pixel in the image. In Figure~\ref{fig:heatmaps}
we show some example predictions shown as heatmaps of the probability of the
smooth shading class (\smooth{}). In the left image, the network correctly predicts
smooth shading on the wall and polished stone surfaces. Of particular
use are high probability predictions on textured surfaces that have
smooth shading, because these are non-trivial to predict based on image
intensity alone.

However, our method also makes some mistakes. In a few cases, the high
probability areas ``bleed over'' shape discontinuities, as in the
corner of the bathroom in the left image, or the sharp shape discontinuities of
the trolley in the right image.
In general, the network predicts smooth shading somewhat conservatively, and misses some smooth
shading regions, but it does well in predicting non-smooth regions in most
cases. Please see the supplemental material for further discussion and hundreds of heatmap predictions.

\section{Evaluation}
Since to our knowledge there are no existing algorithms for explicitly
predicting the three types of shading classes we consider, we focus our evaluation
on predictions of smooth shading vs.\ non-smooth shading, where we can use
simple baselines for comparison.

\subsection{Baselines}
A natural set of baselines for predicting shading categories are
intrinsic image algorithms, which take an input image $I$ and
decompose it into reflectance \ri{} and shading \si{} layers.  We
use several state-of-the-art intrinsic image decomposition algorithms
as baselines.
%%%%%%
In particular, given a decomposition \ri{} and \si{}, we classify a pixel
$p$ as smooth/non-smooth shading based on the gradient magnitude of
the shading channel \si{}. If the gradient magnitude at $p$ is less than
a threshold $\tau$, i.e., $\norm{\nabla \sie{}(p)}_2 < \tau$,
then we say the predicted shading is smooth at $p$ (otherwise,
non-smooth).  In practice, we found that applying a maximum filter of
size $10\times 10$ on the gradient image improved the results of these
baselines, so we apply this filtering in our tests.
%%%%%%%
We compare our CNN predictions to seven baseline algorithms: (1) ``Constant
Reflectance'' (i.e., the shading channel is the luminance channel of
the input image itself), (2) [Shen \etal~2011]~\cite{shen-CVPR2011b}, (3) Color
Retinex~\cite{grosse2009}, (4) [Garces \etal~2012]~\cite{garces2012}, (5) [Zhao
\etal~2012]~\cite{zhao-PAMI2012}, (6) [Bell \etal~2014]~\cite{bell2014intrinsic},
and (7) [Zhou \etal~2015]~\cite{zhou2015learning}.

\begin{figure}[t]
\centering
\includegraphics[width=\linewidth]{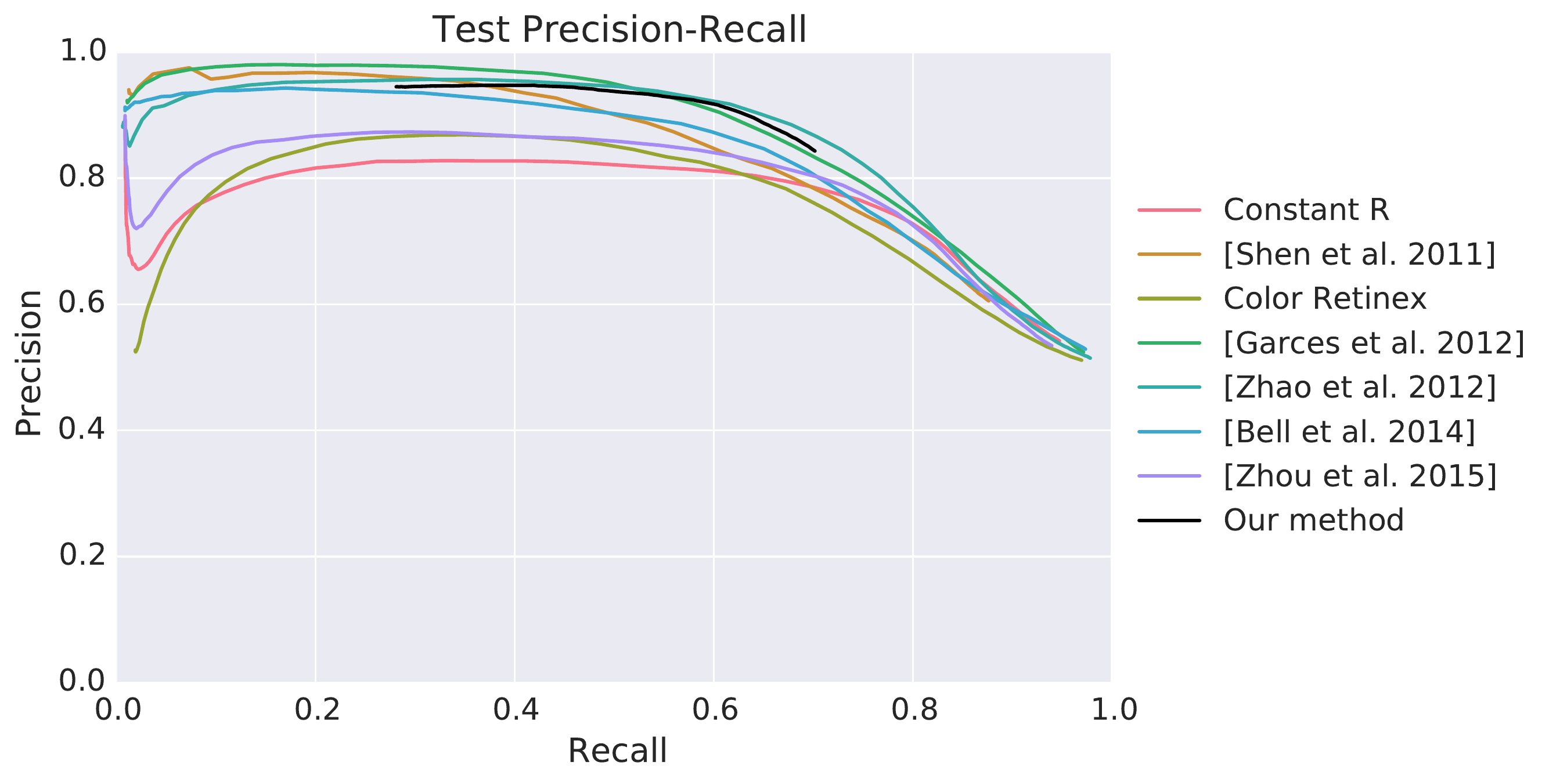}
\caption{{\em Precision-recall for shading predictions}. We plot PR curves for
	the baselines and our algorithm. Our algorithm has competitive performance; see text for discussion.
}
\label{fig:pr}
\end{figure}

\subsection{Precision-Recall}\label{subsec:pr}
By running these baseline algorithms on our test image set, and
sweeping the threshold $\tau$, we can plot precision-recall (PR)
curves for the smooth shading class predicted by the
baselines (see the colored lines
in Fig.~\ref{fig:pr}).\footnote{Note that our test set has the same 2:1:1 label
  balance as the training set, and we only include non-smooth shading
  points exactly on boundaries in our test set, i.e., we do not
  perform label dilation. We do not evaluate on points that have no label in
the dataset.}  Similarly, we can sweep a threshold
$\sigma$ for the smooth shading probabilities predicted by our CNN,
i.e.~we say the shading is smooth at pixel $p$ if the predicted smooth
shading probability $P_p$ is greater than $\sigma$. One way to interpret this evaluation is as a ``smooth shading detector''---the algorithm must classify each pixel as smooth/not-smooth, and we evaluate precision and recall on this classification. The resulting PR
curves are shown in Figure~\ref{fig:pr}, and the performance at several recall values on these curves are shown in Table~\ref{table:pr}.
The best methods are [Garces \etal~2012] and [Zhao \etal~2012], which use global
optimization including clustering and long-range terms.  In comparison, our
method uses a single feed-forward pass and still achieves competitive
performance.

\begin{table}[t]
\centering
\begin{tabular}{c|c|c|c|}
  \multicolumn{1}{c|}{} & \multicolumn{3}{c|}{Precision @}\\
  {\bf Method} & 30\% & 50\% & 70\% \\
  \hline
  \hline
    Constant R & 0.827 & 0.822 & 0.787 \\
  \hline
    [Shen \etal~2011]~\cite{shen-CVPR2011b} & 0.958 & 0.899 & 0.784 \\
  \hline
    Color Retinex~\cite{grosse2009} & 0.867 & 0.850 & 0.755 \\
  \hline
    [Garces \etal~2012]~\cite{garces2012} & {\bf 0.977} & {\bf 0.949} & 0.834 \\
  \hline
    [Zhao \etal~2012]~\cite{zhao-PAMI2012} & 0.956 & 0.945 & {\bf 0.868} \\
  \hline
    [Bell \etal~2014]~\cite{bell2014intrinsic} & 0.936 & 0.902 & 0.802 \\
  \hline
    [Zhou \etal~2015]~\cite{zhou2015learning} & 0.873 & 0.858 & 0.802 \\
  \hline
  {\bf Our method} & 0.946 & 0.938 & 0.845 \\
  \hline
\end{tabular}
\caption{{\em Precision of shading predictions at varying levels of
    recall}.  Precision @ recall levels of 30\%, 50\%, and 70\% are
  shown for the seven baselines and our proposed method.}
\label{table:pr}
\end{table}

Note that the PR curves are not all monotonic (see, for example,
the left part of the PR curves), in part because the
ground truth \nsnd{} labels contain a number of pixels that have very low
contrast (i.e., have small image gradients and are thus very
difficult to classify correctly). Some of these low-contrast pixels
are due to effects like saturated pixels in the input imagery (e.g., a
corner of a wall near a strong light source). On the other hand, these ground truth labels
are based on the Kinect depth images whose quality is not limited by image contrast.
These pixels are in the majority among the pixels classified as smooth shading at low recall
regions of the curves where the threshold $\tau$ for the baselines is small. As
$\tau$, and consequently the recall increases, the proportion of low contrast
pixels decreases and the precision increases for a short segment of the curve.

Finally, note that the curve for our method is
truncated on both ends. This is due to the behavior of the CNN, where
the prediction values it produces (after the final softmax layer) are
often saturated at exactly 0 or 1. That is, there are a number of
pixels where it reports maximal confidence in smooth or non-smooth
shading, and this behavior manifests as truncation of the PR curve. As
a result, our maximum recall is lower than that of other methods. This
behavior suggests that the final softmax layer may be eliminating
some useful dynamic range in the prediction scores.

It is interesting to note that [Bell \etal~2014]
outperforms [Zhou \etal~2015]~\cite{zhou2015learning} on this shading prediction task, even though the latter is
considered to have higher quality intrinsic image
decompositions according to the IIW benchmark score~\cite{bell2014intrinsic}. We conjecture that since the IIW benchmark is based only on reflectance
annotations, errors in the decomposed shading layer
are not sufficiently penalized. Hence, our dataset offers another, complementary lens for evaluating the results of intrinsic image method. An area of future work is to use our annotations in conjunction with the IIW benchmark to devise a new, unified method for evaluating intrinsic image algorithms that considers both reflectance and shading annotations. We show example decompositions from both algorithms as supplemental material.

\section{Application to Intrinsic Images}\label{sec:intrinsic}

\begin{figure*}[t]
  \centering
  \begin{tabular}{cccccc}
    \includegraphics[width=0.14\linewidth]{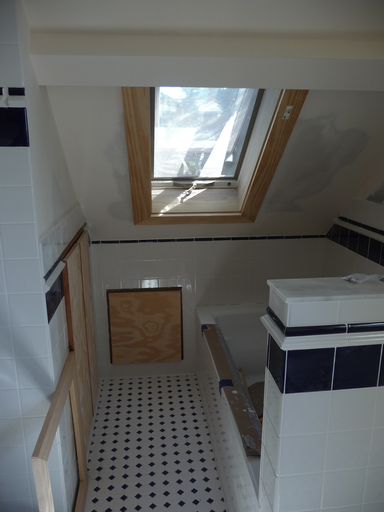} &
    \includegraphics[width=0.14\linewidth]{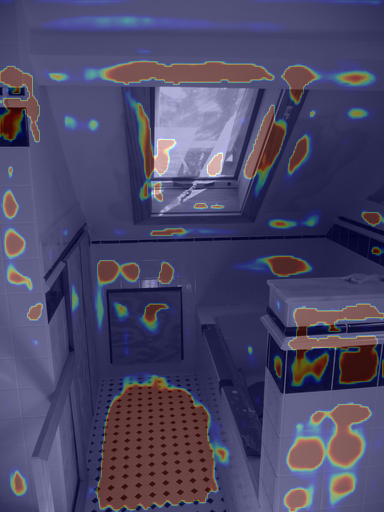} &
    \includegraphics[width=0.14\linewidth]{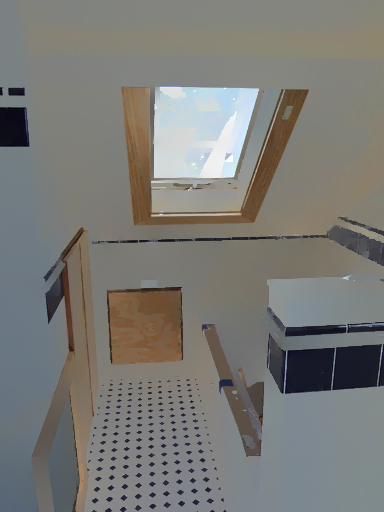} &
    \includegraphics[width=0.14\linewidth]{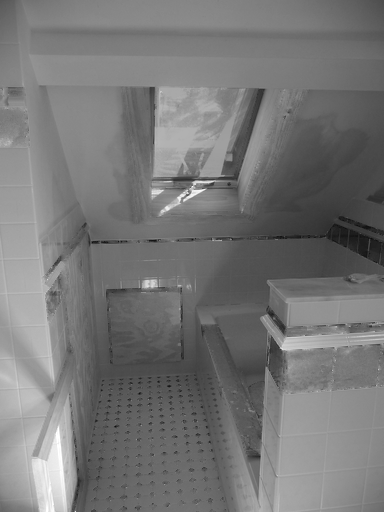} &
    \includegraphics[width=0.14\linewidth]{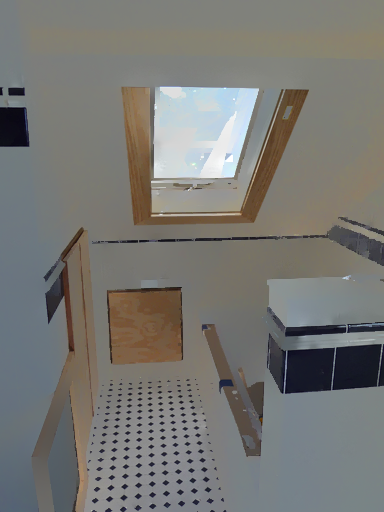} &
    \includegraphics[width=0.14\linewidth]{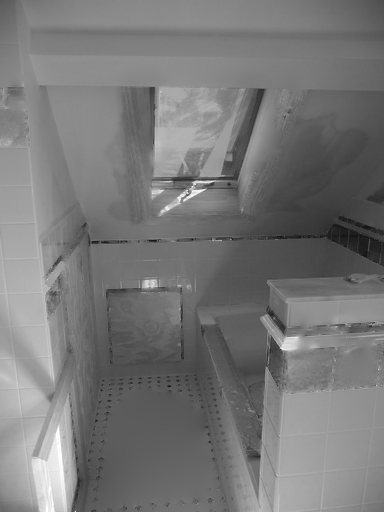} \\
    \includegraphics[width=0.14\linewidth]{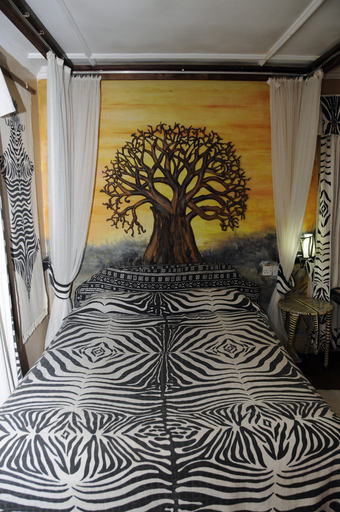} &
    \includegraphics[width=0.14\linewidth]{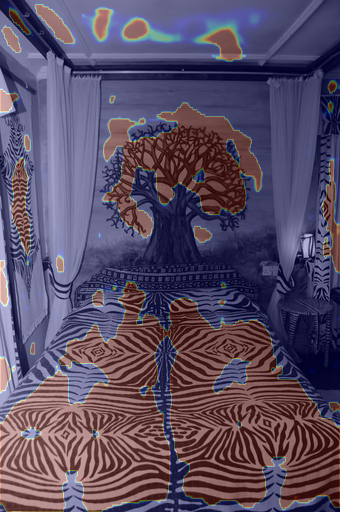} &
    \includegraphics[width=0.14\linewidth]{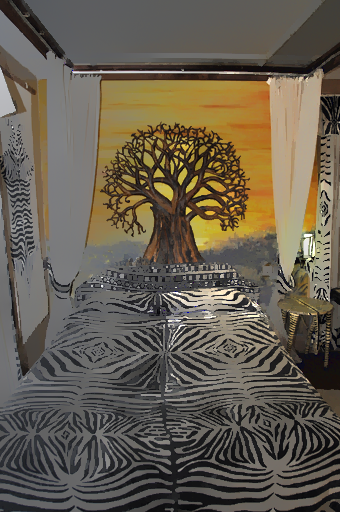} &
    \includegraphics[width=0.14\linewidth]{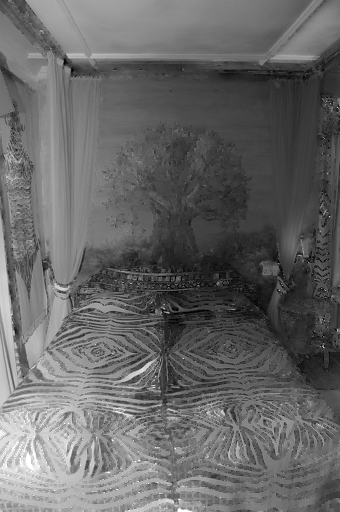} &
    \includegraphics[width=0.14\linewidth]{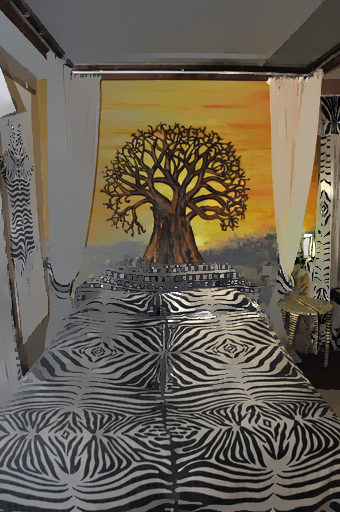} &
    \includegraphics[width=0.14\linewidth]{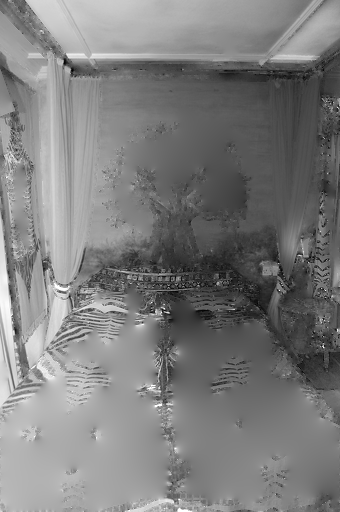} \\
    \small{(a) Input} & \small{(b) Smooth \si{} (\smooth{})} & \small{(c) \ri{}
  w/o prior} & \small{(d) \si{} w/o prior} & \small{(e) \ri{} with prior} &
  \small{(f) \si{} with prior}
  \end{tabular}
\caption{We show intrinsic image decompositions using our smooth shading
  prior. All images are selected from the test set. {\bf (b)} shows a heatmap for our smooth
  shading (\smooth{}) prediction. \ri{} and \si{} denote reflectance and shading respectively. By
  using our smooth shading prior we can reduce artifacts in the decomposed
  shading layer, in particular removing surface texture effects which belong to the
  reflectance layer. For instance, our method removes more of the floor tile texture (top), or the
  texture of the blanket (bottom) from the shading layers.
}
\label{fig:shadingprior}
\end{figure*}

We now demonstrate a use of our smooth shading predictions as a prior for
intrinsic image decomposition algorithms. To demonstrate the use of
this prior, we modified the Retinex formulation of~\cite{zhao-PAMI2012}. The
original cost function they minimize to obtain a decomposition is the following:
%%%%%%%
\[
  f_l(\sie{}) = \sum_{(p, q) \in \mathcal{N}} [(\sie{}_p - \sie{}_q)^2 + \omega_{(p, q)}
    (\rie{}_p -
  \rie{}_q)^2],
\]
%%%%%%
where $\mathcal{N}$ denotes the set of all neighboring pairs of pixels in the
image, $\sie{}_p$ and $\rie{}_p$ are the shading and reflectance at pixel $p$,
respectively.
$\omega_{(p, q)}$ balances between shading and reflectance
smoothness and is determined by the Retinex rule:
%%%%%%%%
\[
  \omega_{(p, q)} = \begin{cases}
    0 & \text{if} \norm{\hat{\rie{}}_p - \hat{\rie{}}_q}_2 > t\\
    100 & \text{otherwise.}
   \end{cases}
  \]
%%%%%%
where $t$ is the Retinex threshold,
%which can be selected by cross-validation,
$\hat{\rie{}}_p$ and
$\hat{\rie{}}_q$ are the chromaticities of pixels $p$ and $q$
(see~\cite{zhao-PAMI2012} for details). For simplicity, we do not use the non-local
constraints of~\cite{zhao-PAMI2012}, only the Retinex constraint with $t =
0.02$.
%%%%%
We incorporate our smooth shading prior by modifying $\omega_{(p, q)}$:
%%%%%
\[
  \omega_{(p, q)} = \begin{cases}
    0 & \text{if} \norm{\hat{\rie{}}_p - \hat{\rie{}}_q}_2 > t\\
    100 \cdot [1 - (H_p + H_q)/2] & \text{otherwise.}
   \end{cases}
\]
%%%%%
$H_p$ and $H_q$ are the smooth shading probabilities predicted by our model
(Section~\ref{sec:heatmap}) at pixels $p$ and $q$. This formulation allows the
decomposition algorithm to smoothly ignore the strong reflectance constancy
constraint at regions where the predicted smooth shading heatmap has high
probabilities.

In Figure~\ref{fig:shadingprior}, we show decompositions with and without our
smooth shading prior. In some cases, we can see significant improvement in the
decomposed shading layer. Specifically, our network is successful in detecting
textured regions with smooth shading where most intrinsic image algorithms fail
to remove the texture from the shading layer. The supplementary has more examples.

\section{Conclusion and Future Work}

We present \dataset{}, a new large-scale dataset of shading in real-world
indoor scenes, created using a combination of crowdsourcing and automation.
Using this dataset, we trained a CNN to achieve competitive performance
against a number of baselines in per-pixel classification of shading effects in
images.  We also demonstrate a potential application of this network as a
smooth shading prior for intrinsic image decomposition. We have made this
dataset publicly available at
\href{http://opensurfaces.cs.cornell.edu/saw/}{http://opensurfaces.cs.cornell.edu/saw}. Illumination is a key property of image formation;
we hope that our dataset can enable other researchers to explore this property
in a richer way that harnesses modern machine learning tools.

Our work suggests a few possibilities for future work.  Evaluation of intrinsic
image algorithms on our data suggests that our annotations may
provide another way to rank these algorithms based on shading performance,
complementary to the widely used WHDR metric~\cite{bell2014intrinsic} that only
directly measures performance on reflectance. Using \datasetshort{}
with the reflectance annotations of IIW, we believe
that new intrinsic image metrics can be established to advance the state
of the art.

Our CNN for classifying pixels into different shading categories could be extended in a number of ways. For instance, we could jointly predict shading categories, shape, and materials (i.e., learn in a multi-task setting inside an approach like PixelNet~\cite{Bansal16}). Pushing this idea further, one could create a network that directly predicts an intrinsic image decomposition along with scene geometry, or further still, one that predicts a full 3D description of geometry and illumination trained using our data.

\medskip
\noindent{\bf Acknowledgment}
This work was supported by the National Science Foundation (grants IIS-1617861,
IIS-1011919, IIS-1161645, IIS-1149393), and by a Google Faculty Research Award.

{\small
\bibliographystyle{ieee}
\bibliography{egbib}
}

\end{document}